\documentclass{article}
\pdfoutput=1
\usepackage{arxiv}
\usepackage{IEEEtrantools}
\usepackage[utf8]{inputenc} 
\usepackage[T1]{fontenc}    
\usepackage{hyperref}       
\usepackage{url}            
\usepackage{booktabs}       
\usepackage{amsfonts}       
\usepackage{nicefrac}       
\usepackage{comment}
\usepackage[pdftex]{graphicx}
\usepackage{epstopdf}
\graphicspath{ {Figs/} }   	
\DeclareGraphicsExtensions{.pdf, .jpeg, .png, .eps}     

\usepackage{amsthm}
\usepackage{amsmath}
\usepackage{amssymb}
\usepackage{algorithm}
\usepackage{algorithmicx}

\usepackage{algpseudocode}
\usepackage{multirow}
\usepackage{color}
\usepackage{enumitem}
\usepackage{cite}

\usepackage{textgreek}
\usepackage{stmaryrd}

\usepackage{arydshln}

\usepackage[caption=false,font=footnotesize]{subfig}

\usepackage{bm}   

\newcommand{\etal}{\textit{et al}. }
\newcommand{\ie}{\textit{i}.\textit{e}. }
\newcommand{\eg}{\textit{e}.\textit{g}.}

\begin{document}
\title{A Flatter Loss for Bias Mitigation in Cross-dataset Facial Age Estimation}

\author{{Ali Akbari, Muhammad Awais, Zhen-Hua Feng, Ammarah~Farooq and Josef Kittler} \\
Centre for Vision, Speech and Signal Processing (CVSSP)\\
University of Surrey, Guildford, UK\\
{\tt\small \{ali.akbari,m.a.rana,j.kittler\}@surrey.ac.uk}
}

\maketitle

\begin{abstract}
The most existing studies in the facial age estimation assume training and test images are captured under similar shooting conditions. 
However, this is rarely valid in real-world applications, where training and test sets usually have different characteristics. 
In this paper, we advocate a cross-dataset protocol for age estimation benchmarking.
In order to improve the cross-dataset age estimation performance, we mitigate the inherent bias caused by the learning algorithm itself.
To this end, we propose a novel loss function that is more effective for neural network training. 
The relative smoothness of the proposed loss function is its advantage with regards to the optimisation process performed by stochastic gradient descent (SGD).
Compared with existing loss functions, the lower gradient of the proposed loss function leads to the convergence of SGD to a better optimum point, and consequently a better generalisation.
The cross-dataset experimental results demonstrate the superiority of the proposed method over the state-of-the-art algorithms in terms of accuracy and generalisation capability.
\end{abstract}

\section{Introduction}
With the emergence of Deep Neural Networks (DNN), the performance of age estimation systems has improved considerably.
However, most of the existing deep-learning-based age estimation methods use the images captured under similar conditions for training and test. This paper focuses on cross-dataset age estimation for practical scenarios, in which the training and test sets usually have different distributions and imaging characteristics.
Unfortunately, the accuracy of age estimation systems can degrade significantly under a cross-dataset setting.
To mitigate this problem, a learning  algorithm should train models which accurately estimate the subject's age under cross-dataset setting.
However, this direction of research has not been adequately addressed by the research community. 

There are many factors, such as gender, race, illumination conditions, image quality, makeup, lifestyle, cosmetic surgery, etc., which may confound the training process~\cite{Akbari20201}.
Since a neural network efficiently learns the data distribution, it is likely to learn the influence of these confounding factors and biases present in the data, instead of learning the actual discriminative cues.
In these situations, the model performs best with test data that is similar to that used to train the model, but drops in accuracy when the test data is collected in unseen conditions.
This motivates us to design models with better generalisation capability.

The cross dataset testing could be formulated as a domain adaptation problem\cite{Jing2019}. 
In this approach, one needs a prior knowledge or a subset of images from the new domain so that a mapping between the original domain and the target domain can be determined. 
However, in many scenarios, such an approach leads to the generalisation of the trained model in a few domains. 
This may not be appropriate for the age estimation problem where the system needs to perform well on any input image from any domain.
It is an expensive task to account for all possible domains.
Beside that, there exists another limitation with domain adaptation approaches. At the test stage, there is no knowledge about the domain which the tested subject falls into and it needs a discriminator to determine which model (original model or domain-shifted model) should be used to estimate the subject's age. 
These studies motivate us to explore a new learning methodology that can render an domain-invariant age estimation solution, exhibiting better generalisation capability than the existing systems.

The design of any machine learning algorithm involves the minimisation of a loss function.
The choice of an inappropriate loss function leads to a systemic bias which appears during the training phase. 
In addition, inadequate and unbalanced training data with respect to the above0mentioned confounding factors can also impact on the learnt solution in terms of a dataset bias.
The existence of systematic and dataset biases result in a solution which might have a poor performance in unseen scenarios.

One way to address the dataset bias is to ensure that there exist roughly the same number of samples from each confounding factor during the training stage.
This may involve heavy data augmentation techniques~\cite{info11020125}, collecting images for significantly underrepresented factors or removing some images from the more dominant factors to re-establish a balance.
However, this can prove to be a challenge as many of the ageing datasets are not annotated with these factors.
It is a cumbersome task to account for all possible factors when curating large-scale datasets or using data augmentation techniques.
Moreover, ageing data is relatively scarce and expensive to generate.
On the other hand, it should be emphasised that if the training algorithm itself exposes the inherent bias, no amount of data collection and augmentation leads to a good generalisation performance of the learnt model.

In order to improve the generalisation performance, we need to mitigate the dataset bias together with the systemic bias.
To this end, we propose a new loss function in this paper to mitigate the systematic bias. 
Our work is motivated by the recent study, which argues that a smoother loss function leads to better generalisation~\cite{Xu2012,Seong2018TowardsFL}.
The relative flatness of the proposed loss function controls the inherent bias of the learnt age estimation model.
Its lower gradient facilitates the discovery of and convergence to a better optimum, and consequently a better generalisation.
Consequently, the better generalisation capability is achieved and the higher accuracy is obtained when the model is tested in unseen (cross-dataset) scenarios.
We stress that developing a useful machine learning model, being robust to the both systemic and dataset biases, is left as a future work.

Finally, in this paper, we introduce a novel \textit{subject-exclusive cross-dataset (SC)} protocol for reliably evaluating the performance of the age estimation systems.
We evaluate the proposed loss function on several well-known benchmarking datasets using the proposed SC protocol. The experiments demonstrate that the proposed loss function defines the state-of-the-art results on this protocol.

\section{Related Work}
With the success of novel machine learning methods, and specifically deep  learning, in a wide variety of fields~\cite{FatemifarICIP2020,Angulu201842,wingloss,Akbari2559,Akbari20176,Akbari20171,KhalidBHIS2020410,Bashar20201678,Bashar20201,Akbari20161},
the research focus has shifted towards learning discriminative features by training a deep neural network end-to-end on labelled datasets~\cite{Carletti2019}.  
In the age estimation area, these end-to-end approaches can be categorised into four major groups, namely regression, ranking, classification and label distribution based methods.
Regression-based algorithms~\cite{Han20151148,Guo2009112,Huang20174058} make the problem definition intuitive by treating age labels as real and continuous values.
The classification-based methods~\cite{Eidinger20142170,Rothe2018144,Zhang201722492,Liu2018292,Pan20185285} cast age estimation as a multi-class classification problem. 
To consider the ordinal relationship among age labels, ranking-based methods~\cite{Chang2011585,Chen2017742,Zeng20191906} and label distribution learning-based methods~\cite{Geng20161734,Gao20172825,He20173849,Shen2017834,Shen20182304,Gao2018712} have been proposed.
The former employ the information about the relative order between ages. 
The latter make use of the correlation among adjacent ages by adopting the concept of label distribution.
These  two  formulations  are at the core of many  methods  in  the  literature.
A recent study~\cite{Gao2018712} proves that the ranking and label distribution learning based algorithms have a linear relationship and so their age estimation performance are very similar.
However, the label distribution learning based methods are more stable during training. 
In the existing label distribution learning based methods, the KL divergence is widely employed as the loss function to measure the similarity between the estimated age distribution and groundtruth~\cite{He20173849,Gao2018712}. 
Most these studies adopt the intra-dataset evaluation to show the performance of the proposed methods. 
Different from the exiting work, we propose a novel loss function and evaluate our system under the cross-dataset setting which is more challenging.

\section{The Proposed Method}
In this section, we first analyse the limitations of the widely-used loss functions in the DNN-based age estimation. 
Then we present a new distribution cognisant loss and discuss the characteristics of the proposed loss.
\subsection{Problem Formulation}
Let $\mathcal{S} = \{(\textbf{x}_n, y_n), n={1,\cdots, N}\}$ denote a set of training samples, where $\textbf{x}_n$ represents the $n$-th training image and $y_n \in [1, L]$ is the corresponding age label.
The ultimate goal of an age estimation algorithm is to discover a mapping function from the input space (images) to the output space (age label).
To this end, a typical method is the one-hot strategy in which a binary vector 
$\textbf{s}=[s_1, \cdots, s_L]^T \in \mathbb{R}^L$ is assigned to the age label.  
If a face image belongs to the $i$-th age label, the $i$-th element in $\textbf{s}$, \ie $s_i$ equals 1, otherwise $s_i=0$. 
This kind of label encoding models the age estimation problem as a standard classification problem.

\begin{figure}[!t]
\centering
\includegraphics[trim={0mm 20mm 0mm 0mm},clip, width=0.25\linewidth]{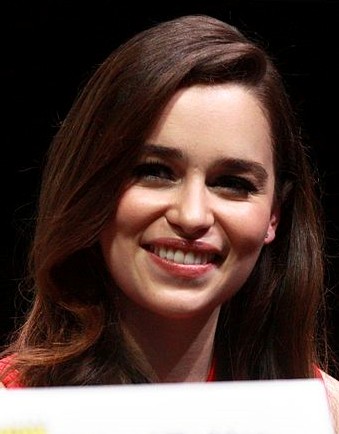}
\includegraphics[trim={0mm 5mm 0mm 0mm},clip, width=0.4\linewidth]{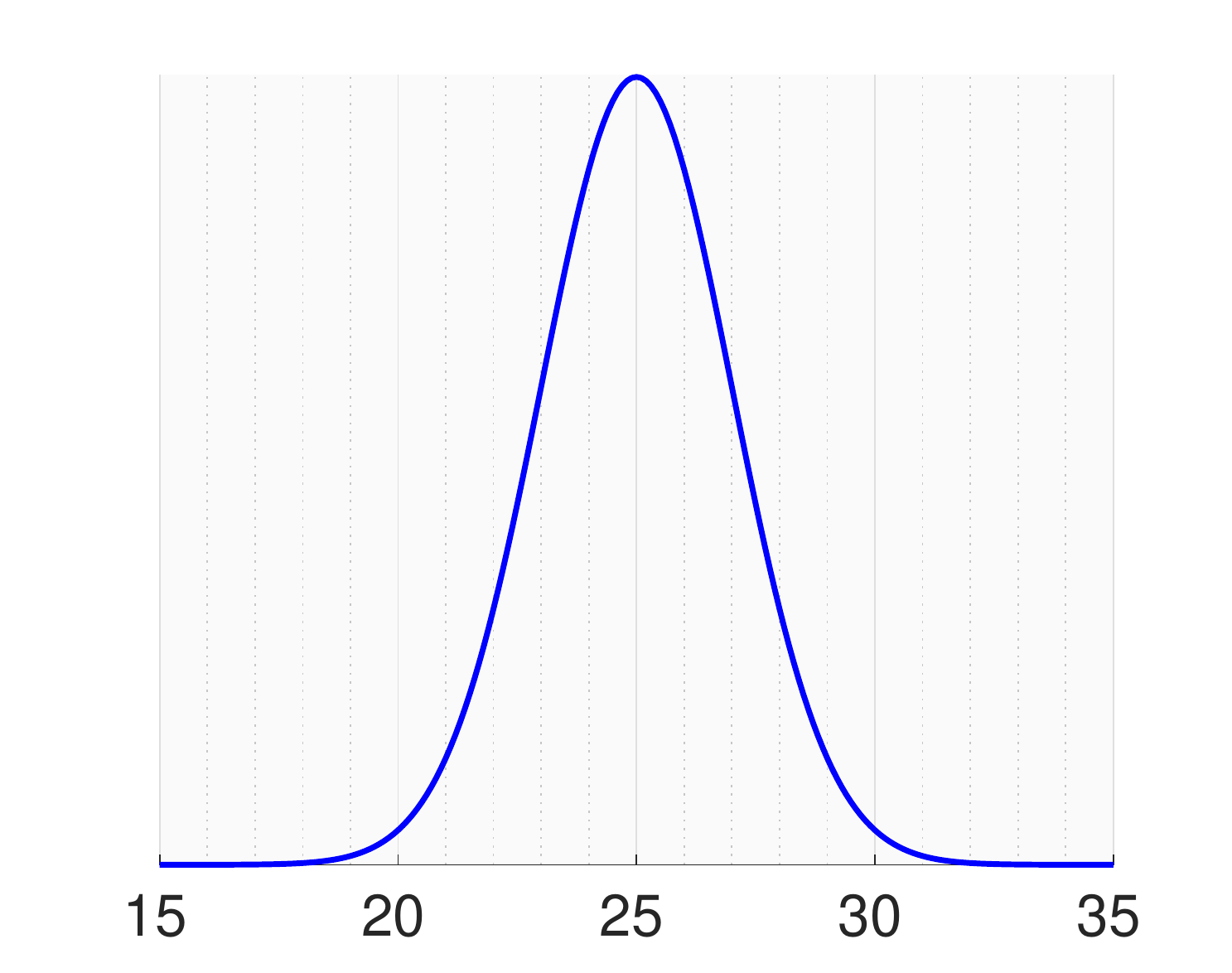}
\caption{Facial age estimation as label distribution learning problem -- due to the similarity between neighbouring ages, the scalar age label is encoded a Gaussian label distribution for a facial image at the age of $25$.}
\label{fig:applications}
\end{figure}

There is a strong visual semantic correlation among the face images of nearby ages, \eg a person at age 30, 31 and 32 may have very similar facial features that are difficult to distinguish. 
By formulating the age estimation problem as a classification problem, this semantic correlation is ignored, leading to an inconsistency during DNN training. 
Therefore, a network trained with the one-hot age labels has difficulty to separate visually similar face samples that have different age labels. 
This problem has been partially addressed by using \textit{label distribution} concept~\cite{Geng20161734}, in which a set of degrees $\textbf{q}=[q_1, \cdots, q_L]^T\in\mathbb{R}^L$ 
is assigned to each label $y$, where $q_i\in [0,1]$ and 
$\sum_{i=1}^L q_i=1$.
Each $q_i$ can be considered as the probability that a face sample $\textbf{x}$ is associated with the age label $i$. 
Fig.~\ref{fig:applications} shows an instance at age $25$ and its corresponding label distribution.

Now, the goal is to learn the mapping function that projects a face image to its corresponding label distribution $\textbf{q}$.
In this paper, we follow the distribution learning approach and, similar to~\cite{Gao20172825,Geng20161734}, a Gaussian distribution, centred at $y$ with a standard deviation $\sigma$, is considered for each label, where the shape (width) of the distribution at each age is controlled by $\sigma$. 
We consider a constant value of $\sigma=2$ at all ages.

\subsection{Analysis of Loss Functions for Age Estimation}\label{sec:analysis}
The Cross Entropy (CE) loss function and Kullback–Leibler (KL) divergence have been widely adopted for training DNN-based models. The former uses one-hot labels and later adopts label distributions. 
In this section, we analytically discuss the limitations of these loss functions in DNN-based age estimation. 

Let us consider a DNN model with parameters $\theta$. Given an input $\mathbf{x}$, its output is $\mathbf{z} = f^\theta(\mathbf{x})$.
Suppose the last layer of the model is followed by a softmax layer
to collapse $\mathbf{z}$ into a probability distribution vector $\mathbf{p}$, where $p_i=\exp(z_i)/\sum_k\exp(z_k)$ denotes the probability of sample $\mathbf{x}$ belonging to class $i$.

The CE loss is the most widely-used loss function for the multi-class classification problem.
When it is applied to the age estimation problem, the CE loss considers the ages as independent classes~\cite{Rothe2018144}.
Consequently, the model ignores the similarity among the neighbouring ages. 
To tackle this issue, Pan~\etal~\cite{Pan20185285} propose to use the CE-MV loss function:
\begin{equation}
\label{ICCV2019:CE-MV_loss}
L_{CE-MV}(\textbf{p}, y)= -\log p_k + \lambda_1(\mu_p - y)^2 + \lambda_2 \sigma_p^2,%
\end{equation}
where the CE loss is regulated with the mean and variance (MV). $\lambda_1$ and $\lambda_2$ are regularisation parameters,
$\mu_p$ and  $\sigma_p$ denote the mean and standard deviation of the predicted distribution.
The second term (mean) punishes the difference between the mean of the predicted and groundtruth label distributions, \ie $\mu_p$ and $y$ respectively. 
The third term (variance) penalised the standard deviation of the predicted label distribution, $\sigma_p$. 
The main concern with CE-MV is its instability caused by outliers.
In fact, due to sensitivity of the mean and variance terms to outliers, large errors with very large gradients are produced during training which may affect the convergence of the network.

The KL loss is the widely-used loss function within the label distribution learning framework~\cite{Gao20172825,Geng20132401} defined as:
\begin{equation}
\label{ICCV2019:KL}
\small
L_{KL}(\textbf{p}, \textbf{q})=\sum\nolimits_k q_k\log (q_k / p_k),%
\end{equation}
where $\textbf{p}$ and $\textbf{q}$ are the predicted and groundtruth distributions.
The KL loss takes positive values between $0$ and $\infty$; the lower the KL value, the better $\textbf{p}$ matches $\textbf{q}$. 

\begin{figure}[!t]
\centering
\begin{minipage}[b]{\linewidth}
\centering
\subfloat[]{\includegraphics[width=0.33
\linewidth]{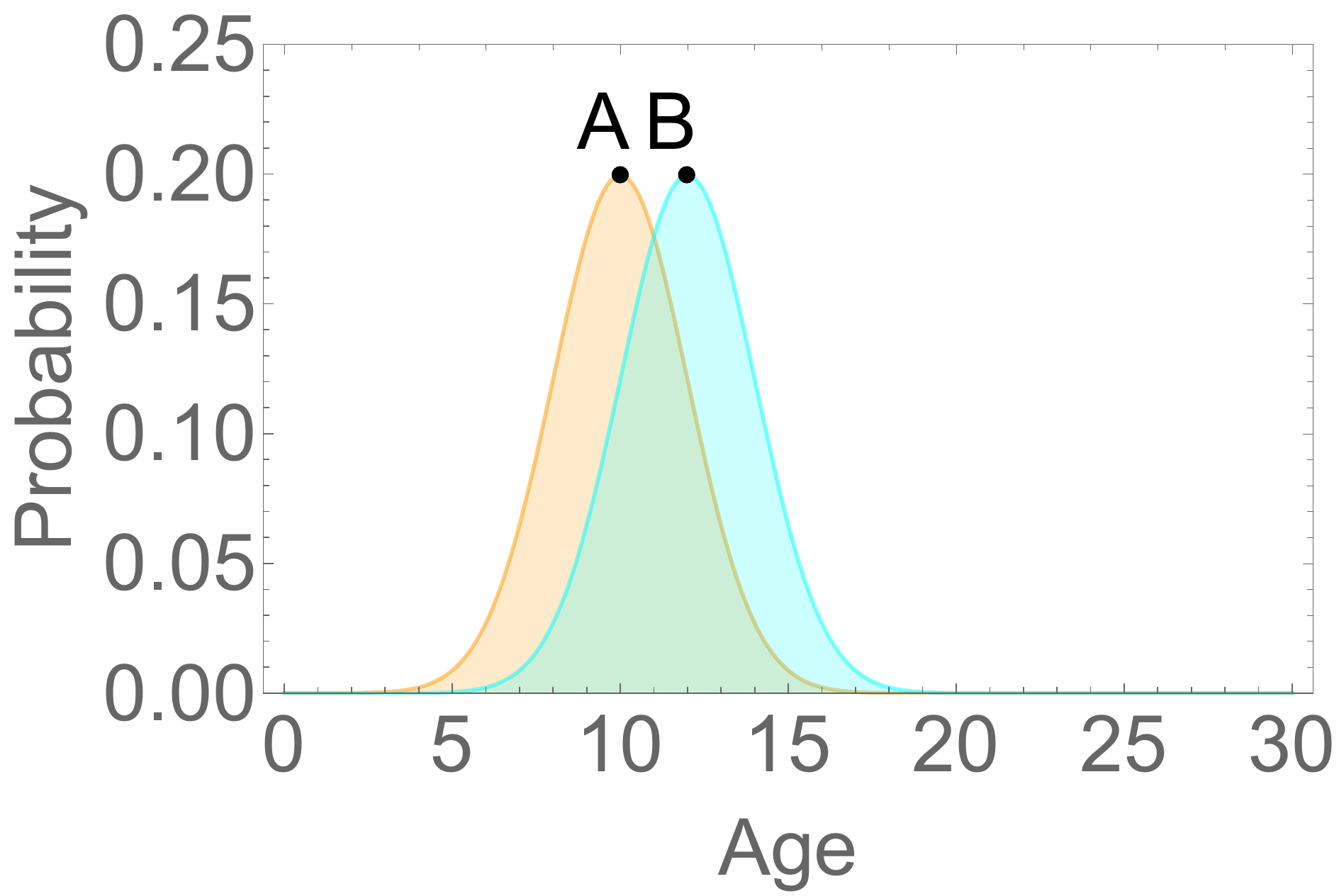}%
\label{fig:Gaussians}}
\subfloat[]{\includegraphics[width=0.33
\linewidth]{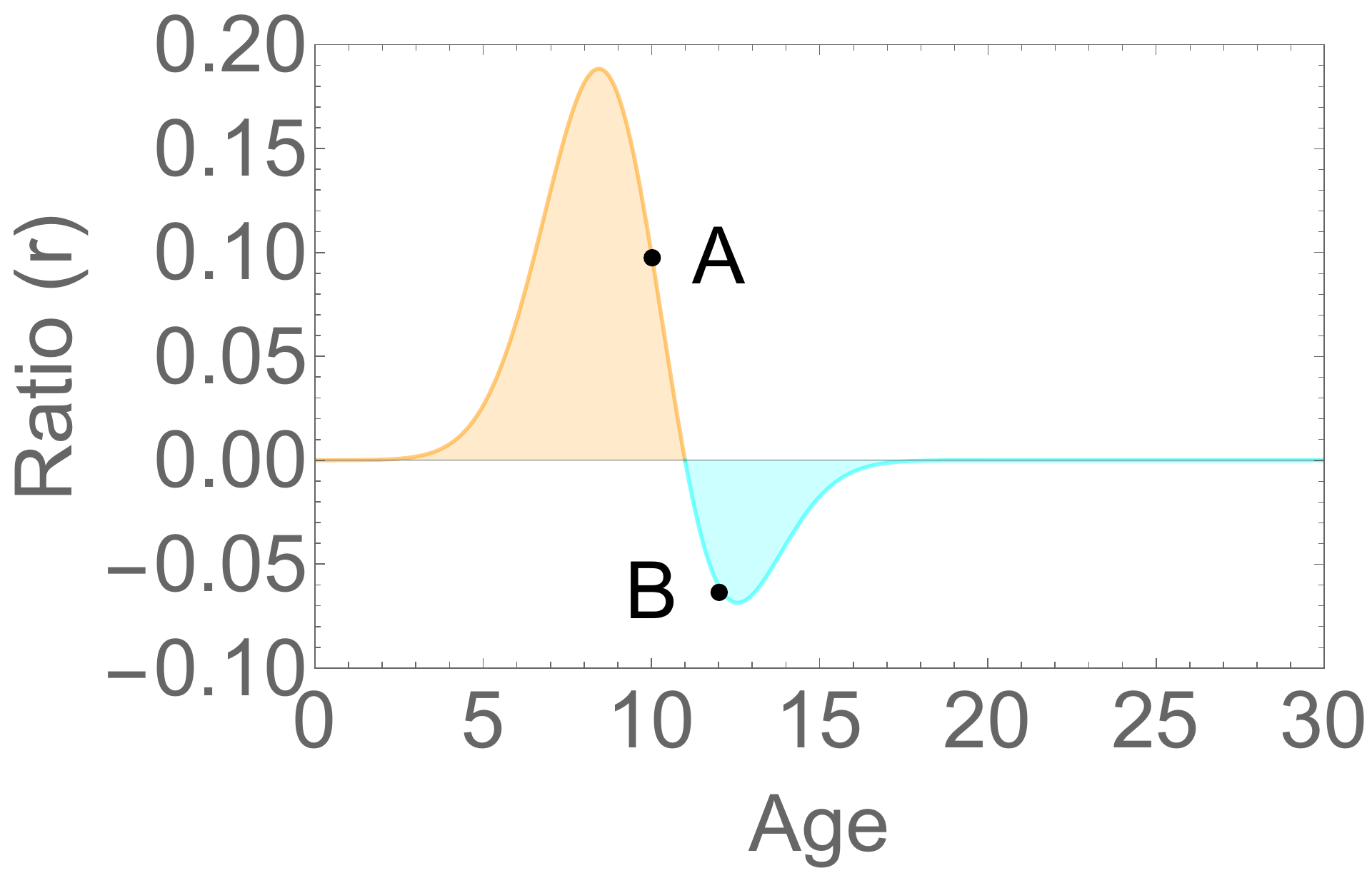}%
\label{fig:KL}}

\subfloat[]{\includegraphics[width=0.33\linewidth]{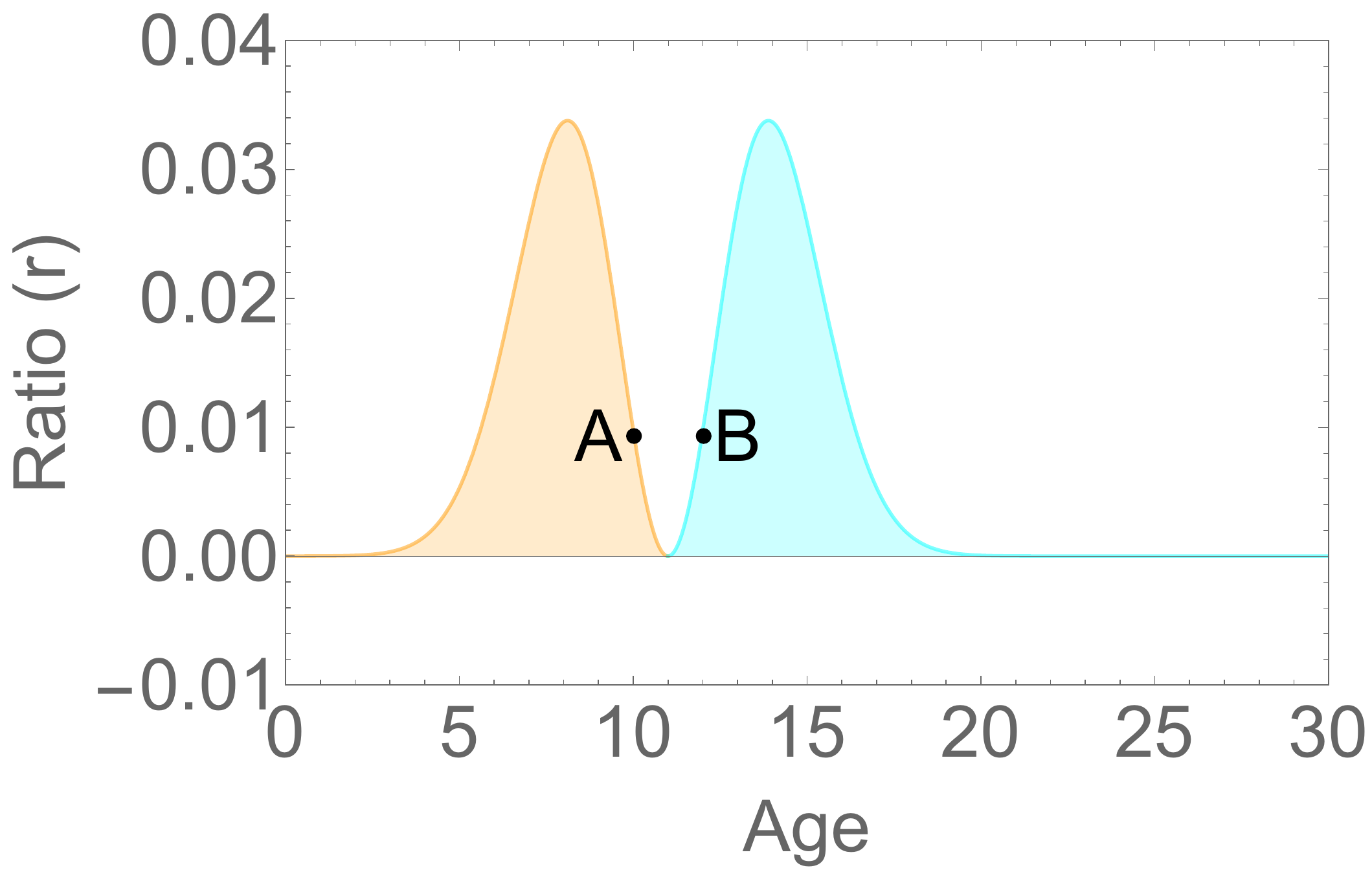}%
\label{fig:MD}}
\caption{Comparison of the KL and proposed loss functions between two Gaussian distributions. \protect\subref{fig:Gaussians} Two Gaussian distributions. \protect\subref{fig:KL} KL loss function. \protect\subref{fig:MD} proposed loss function.}
\label{fig:assymetric}
\end{minipage}
\end{figure}

One concern with the KL loss is its asymmetric nature, \ie $L_{KL}(\textbf{p}, \textbf{q}) \neq L_{KL}(\textbf{q}, \textbf{p})$. 
If $q_i$ is greater than $p_i$, the ratio $r_i=q_i\log(q_i/p_i)$ generates positive values. 
Inversely, if $q_i$ is smaller than $p_i$, $r_i$ produces negative values. 
As shown in Fig.~\ref{fig:KL}, this leads to non-uniform matching over the age range.
For example, consider points $A$ and $B$ in Fig.~\ref{fig:assymetric}.
Two distributions have equal distance at these points, but the values of $r_i$ are different.
Therefore, when $q_i$ is greater than $p_i$, the contribution to the total risk will be more significant. 
That is, the point $A$ contributes more to the total risk than the point $B$. 
After minimising the distance using the KL loss, a better match of $p_i$ to $q_i$ is expected where $q_i$ is greater than $p_i$. 

Another concern with of the KL loss is related to the update rule of model's weights in the back-propagation step.  The derivative of the KL loss is obtained as:
$\partial L_{KL} / \partial z_i = p_i - q_i$.
So, the update of model's weights obtains by the difference of the corresponding bins on the predicted and groundtruth distribution, \ie $p_i-q_i$, irrespective of the contribution of the adjacent bins.
This might renders the learning process less robust.

\subsection{The Proposed Loss Function}
In this paper, we proposed a new loss function, distribution cognisant loss, for age estimation. We define the proposed loss as:
\begin{equation}
\label{ICCV2019:MDloss}
\small
L_{DC}= \log(1 - \alpha(1 - \sum\nolimits_{k=1}^L\sqrt{p_k q_k})) / \log(1 - \alpha)
\end{equation}
where $\alpha$ is the loss parameter satisfying $0 < \alpha < 1$. 
As will be discussed later, the best performance is achieved when $\alpha$ gets a very small value near to $0$\footnote{An alternative formulation for the proposed loss is proposed in our recent work~\cite{AkbariPami20}.
It provides more desirable properties and achieves nearly the same performance with Eq.~\eqref{ICCV2019:MDloss}. Interested readers may consider the proposed formula in~\cite{AkbariPami20} as the loss function for training DNN models.}.
The proposed loss function avoids the above-mentioned concerns associated with the CE and KL loss functions.
Therefore, it can be considered as a more robust measure for obtaining the difference between two label distributions.
In the following, we discuss the desirable properties of proposed loss function.

\begin{itemize}
    \item The proposed loss has a symmetric shape and is bounded in the range $[0,1]$ for any value of $\alpha$. Due to these characteristics, as illustrated in Fig.~\ref{fig:MD}, the proposed loss exhibits uniform matching over the whole age range.
    \item Our proposed loss tackles the singularity issue that occurs when $p_i$ is a non-zero value and $q_i \to 0$.
    \item The derivative of our proposed loss with respect to $\textbf{z}_i$ is $\small
\partial L_{DC} / \partial z_i
=
K(\sqrt{p_iq_i}-p_i\sum\nolimits_{k=1}^L\sqrt{p_k q_k})
/ 
(1 - \alpha(1 - \sum\nolimits_{k=1}^L\sqrt{p_k q_k}))$
,
where $K=\alpha \log e / 2\log(1 - \alpha)$.
As can be inferred, the update rule of the model's weights depends on difference between all corresponding bins of two distributions.
This is more robust in comparison with the KL loss function. 
\end{itemize}

Thanks to its desirable ability to limit the systemic bias of the learnt solution, the proposed loss function provides a good performance in unseen scenarios. 
Note that the loss function controls the search space resulting in a model which is perfect for all test sets. 
However, we are also interested in the reason of its generalisation performance. 
We start by presenting an analysis of the generalisation performance of DNN to show why the proposed loss function mitigate the systematic bias. Focusing on the generalisation error defined in~\cite{Xu2012,Seong2018TowardsFL}, we demonstrate that our proposed loss function exhibits improved  generalisation.

Note that we aim to learn a model $f^\theta:\mathcal{X} \rightarrow \mathcal{Q}$, characterised by  $\theta\in \mathcal{H}$, between the input space $\mathcal{X}$ and its related output space $\mathcal{Q}$. 
A common setting of this learning process is given by
\begin{equation}
\label{Eq:popoltion_risk}
\operatorname*{argmin}_{\theta\in \mathcal{H}} \mathbb{E}_{\mathbf{s}\sim\mathcal{D}} [\ell(f^\theta; \mathbf{s})], 
\end{equation}
which produces  model $f^\theta$ over the hypothesis space $\mathcal{F}$ minimising the true risk $R_{\mathrm{true}}(f^\theta)\triangleq\mathbb{E}_{\mathbf{s}\sim\mathcal{D}}[\ell(f^\theta; \mathbf{s})]$ w.r.t. sample $\mathbf{s}=(\mathbf{x}, \mathbf{q})\in \mathcal{X}\times \mathcal{Q}$, defined for an unknown distribution $\mathcal{D}$.
Note that the term $\ell: \mathcal{Q}\times \mathcal{Q} \to \mathbb{R}^+$ presents the loss function and  is used to investigate the accuracy of a hypothesis $f^\theta$ using discrepancy between the predicted and real outputs.

As we assume that the distribution $\mathcal{D}$ is not known, it is not possible to directly solve the optimisation problem~\eqref{Eq:popoltion_risk}. 
Hence, we estimate $R_\mathrm{true}(f^\theta)$ using an empirical risk over the training set. 
Let us consider a finite set of $N$ training samples $\mathcal{S}=\{\mathbf{s}_n, n={1,2,\cdots, N}\}$, where $\mathbf{s}_n=(\mathbf{x}_n, \mathbf{q}_n)$, i.i.d. samples taken from unknown distribution $\mathcal{D}$. The term  $R_\mathrm{emp}(f^\theta) \triangleq\frac{1}{N} \sum_{n=1}^N \ell(f^\theta; \mathbf{z}_n)$ presents the  empirical risk.

Having the training set $\mathcal{S}$, the generalisation error of $f^\theta_{\mathcal{S}}$, trained by exploiting the learning algorithm $\mathcal{A}$ on $\mathcal{S}$, measures the difference between the empirical and true risk, \ie $E(\mathcal{f^\theta_{\mathcal{S}}}) = R_\mathrm{true}(f^\theta_{\mathcal{S}}) - R_\mathrm{emp}(f^\theta_{\mathcal{S}})$.

Seong \etal~\cite{Seong2018TowardsFL} demonstrate that the generalisation error is upper-bounded by a function of the loss function's smoothness w.r.t. changes on the loss surface.
This result 
could be useful for analysing  the performance of neural networks w.r.t the employed loss function. 
The main message is:
\textit{ the smoother loss surface provides a better generalisation for the output model trained by that loss function.}

\begin{figure}[t]
\centering
\includegraphics[width=.7\linewidth]{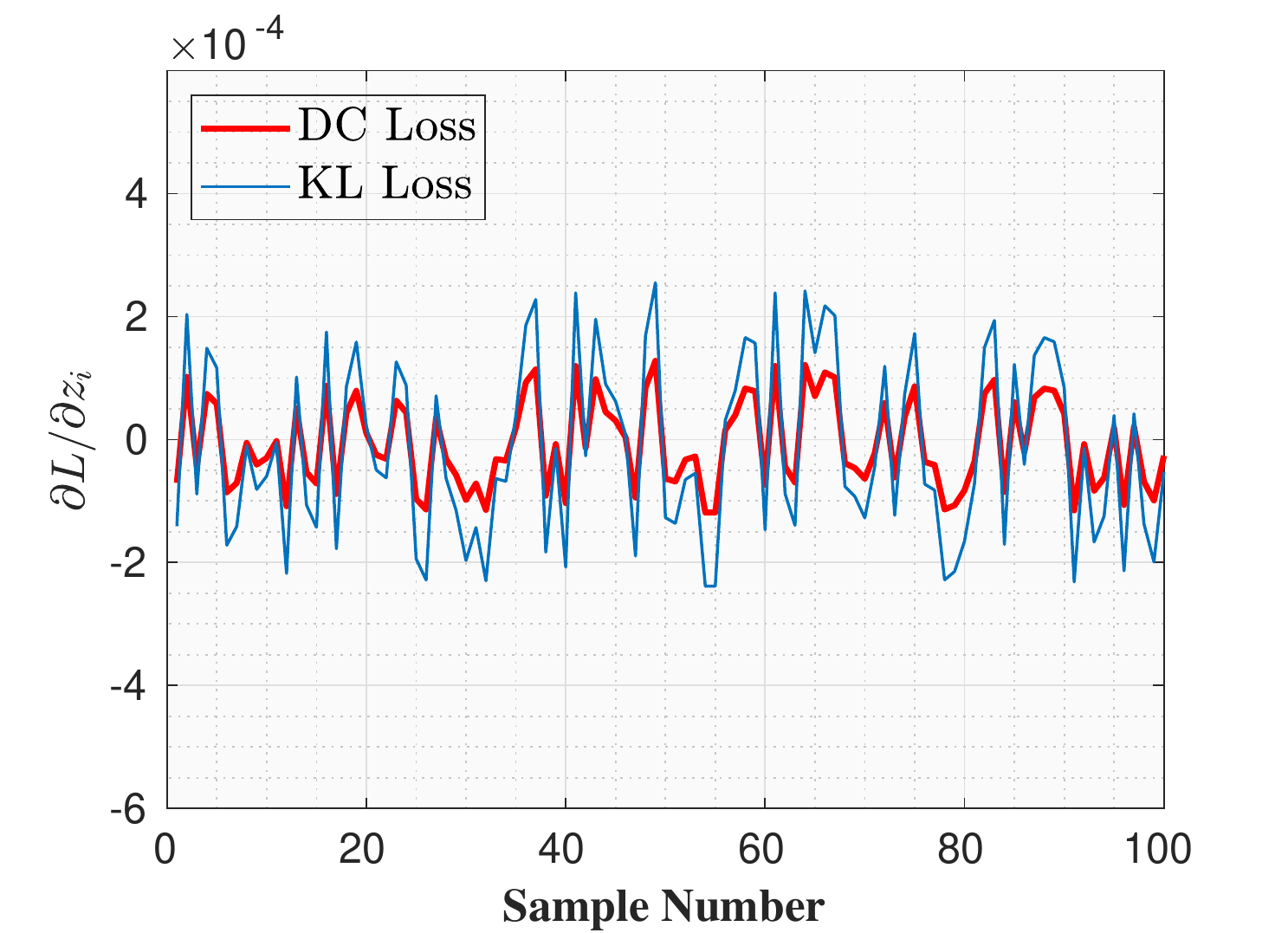}
\caption{Behaviour of gradient of the KL loss and the proposed DC loss. 
}
\label{fig:Derivatives}
\end{figure}

The important result of the theoretical analysis is that by carefully selecting the loss function, one is able to get a better bound on the generalisation error.
To see the significant properties of the proposed loss function, consider a groundtruth distribution labelling the age of 50.
Through randomly shifting the groundtruth distribution and then adding a uniform noise to the shifted distributions, we generate $100$ distributions.
We consider these distributions as generating the estimated output label distribution after the softmax layer.
Fig.~\ref{fig:Derivatives} presents  the gradient of the proposed and KL loss functions for the $100$ samples w.r.t. the $i$-th output of the last FC layer \ie $\frac{\partial L}{\partial z_i}$.
As the figure shows, the magnitude of the gradient of the proposed loss function is always less than that of the KL divergence.
This shows that a better bound on the robustness is achieved by the proposed  loss function.
Hence, it is less steep, compared to that of  the KL loss function, improving the discovery and convergence to a finer optimum, and thus producing better generalisation.  

\section{Subject-exclusive Cross-dataset (SC) Protocol}
In the exiting age estimation protocols, images in the test set have the same shooting characteristics with those in the training set.
For example, the widely-used random splitting protocol adopts $80\%$ and $20\%$ images of a dataset as the training and test sets.
Such evaluation protocols are  likely to provide unreliable and meaningless information about the performance of the trained model in unseen scenarios, where the testing and training images come form different distributions and have distinct shooting characteristics.

In this paper, a new protocol, called \textit{subject-exclusive cross-dataset (SC) protocol}, is proposed which provides more meaningful information for evaluating the age prediction accuracy.
Different from the existing protocols, in our proposed SC protocol, the test set is chosen from another dataset which has completely different shooting characteristics.
That is images from the test dataset should not be utilised during training phase.
Therefore, the trained model is not biased to the test set.
Furthermore, since each person has different ageing process, we guarantee that no image from the same subject exists in the training and testing sets. 
Under these two conditions, the generalisation power of an age estimation system can be accessed more effectively and reliably.
To the best of our knowledge, this the first work to evaluate the performance of several age prediction systems under the SC protocol.

\section{Experimental Results}
\subsection{Settings and Datasets}
We implemented the proposed loss using Matlab powered by MatConvNet~\cite{Matconvnet}. 
Mean Absolute Error (MAE) and Cumulative Score (CS)~\cite{Guo2009112} are adopted for evaluating the age estimation performance. 
The MAE is obtained as $\sum_k \frac{|\hat{y}_k - y_k|}{K}$, where $K$ is the total number of test images. 
The age prediction accuracy is obtained by the CS measure defined as $\frac{K_I}{K}\times 100 \%$, where $K_I$ is the number of the images whose $|\hat{y}_k-y_k|<I$. In this paper, $I$ is set as $5$. 

We use the VGG model, being trained on the facial datasets~\cite{Parkhi2015}, as the backbone of our age estimation system. 
The batch size, momentum and weight decay are set to 80, 0.9 and 0.0005. 
The learning rate decreases exponentially (with the exponential growth $-1$) for 30 epochs from 0.001 to 0.00001. 
All images are aligned with respect to five facial landmarks (eyes centre, nose tip and mouth corners) extracted by the MT-CNN face detector~\cite{MTCNN}. 
The face image is then extracted and resized to $256\times 256$ pixels. 
Standard data augmentation techniques (random flipping, cropping and colour jittering) are employed during the training phase.
At the inference step, the central-cropped image is used as the input of the network and the subject's age is predicted by taking the most probable value of the network's output, as $\hat{y}=\operatorname*{argmax}_i p_i$.

\begin{figure}
\centering
\includegraphics[width=0.7\linewidth]{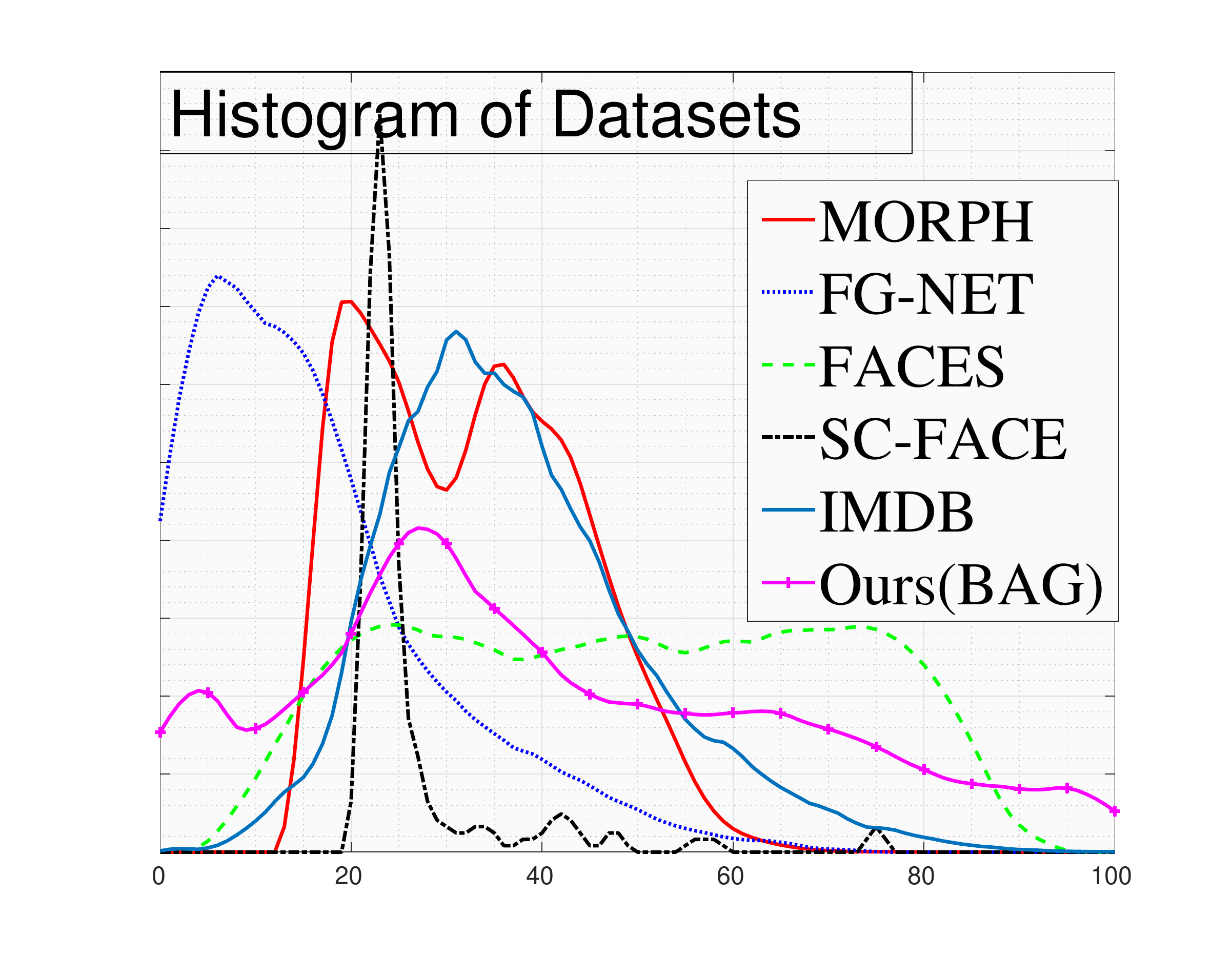}
\caption{Histogram of different databases over the age 0 to 100. }\label{fig:ageEstimationDatasetHist}
\end{figure}

We crawled $23,876$ images of subjects who have ages in excess of $70$ years old and younger than $20$ years old. 
These images were crawled from Internet by using the years as keywords on the Google Images. 
The images in the dataset are then carefully checked by a human annotator to remove irrelevant images. each image. 
This dataset is then combined with the two other existing facial datasets, namely AgeDB and UTKFace\footnote{The AgeDB\cite{Moschoglou20171997} and UTKFace~\cite{Zhang20174352} datasets include $16,488$ and $21,374$ facial images, respectively.} to build a novel dataset for age estimation.
Despite the existing ageing datasets, this new dataset has enough images across different ages, specifically for ages over 70 years-old or younger than 20 years-old. 
The histogram of different databases with respect to the number of images per each age are shown in Fig.~\ref{fig:ageEstimationDatasetHist}.
As seen from, FACES database may look a bit more balanced but it has only $2052$ samples. 
Moreover, the ages between 0 and 19 and 80 and 100 are not present at all\footnote{Our BAG database and trained models are available in https://github.com/Alien84/Cross-Database-Age-Estimation/}.

\begin{figure}[!t]
\centering
\includegraphics[width=0.7\linewidth]{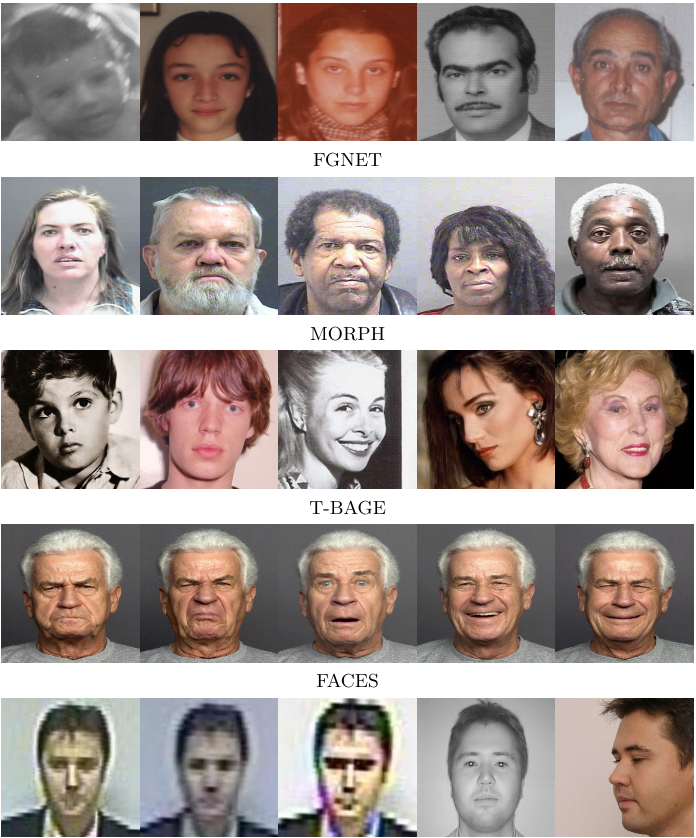}%
\vspace{-1em}
\caption{Sample images from the evaluation databases.}\label{DtatabesesSamples}
\end{figure}

Four benchmarking datasets, including MORPH~\cite{MORPH}, FG-NET~\cite{FGNET}, FACES~\cite{Ebner2010} and SC-FACE~\cite{Grgic2011} datasets, are used as the test sets.
Fig.~\ref{DtatabesesSamples} shows some images of each dataset.
MORPH contains $55,134$ images from different races in the age range from $16$ to $77$ years old. The FG-NET dataset includes $1,002$ images with the large variation in lighting conditions ranging from $0$ to $69$ years old. The FACES dataset has $2052$ images with different expressions in the age range from $19$ to $80$ years old. The SC-FACE dataset contains $2,990$ images captured using one high resolution, one infrared camera and five surveillance cameras. The ages range between $21$ and $75$ years old.

\begin{table*}[!t]
\centering
\caption{Age Estimation Performance (MAE \& CS measures) Under the SC protocol.}\label{tab:openset}
\begin{tabular}{l|cc|cc|cc|cc|cc}
\hline
 & \multicolumn{2}{|c|}{FG-NET} & \multicolumn{2}{|c|}{MORPH} & \multicolumn{2}{|c|}{FACES} & \multicolumn{2}{|c|}{SC-FACE} & \multicolumn{2}{|c}{Average} \\
 \hline
 Method & MAE & CS(\%) & MAE & CS(\%) & MAE & CS(\%) & MAE & CS(\%) & MAE & CS(\%) \\
 \hline
Human~\cite{Han20151148} & 4.70 & 69.5 & 6.30 & 51.0 & NA & NA & NA & NA & 5.50 & 60.25\\
Microsoft & 6.20 & 53.80 & 6.59 & 46.00 & - & - & - & - & 6.39 & 49.90 \\
CE~\cite{Rothe2018144} & 3.57 & 78.94 & 6.54 & 53.38 & 6.59 & 50.83 & 6.19 & 65.05 & 5.86 & 59.50 \\
AGEn~\cite{Tan20182610} & 3.53 & 79.78 & 6.40 & 53.97 & 6.34 & 52.40 & 6.12 & 65.21 & 5.72 & 60.60 \\
KL~\cite{Gao20172825} & 3.24 & 81.54 & 6.01 & 57.36 & 6.11 & 55.60 & 6.52 & 60.64 & 5.55 & 61.98 \\
CE-MV~\cite{Pan20185285} & 3.34 & 80.44 & 6.22 & 55.60 & 6.25 & 54.63 & 6.23 & 64.38 & 5.62 & 61.84 \\
DLDL-v2~\cite{Gao2018712} & 3.35 & 81.44 & 5.80 & 57.30 & {5.92} & {56.68} & 6.52 & 61.61 & 5.48 & 62.77\\
\hline
\textbf{Proposed} & \textbf{3.26} & \textbf{81.57} & \textbf{5.69} & \textbf{58.83} & \textbf{5.92} & \textbf{57.45} & \textbf{5.41} & \textbf{67.90} & \textbf{5.07} & \textbf{66.43}\\
\hline
\end{tabular}
\end{table*}

\subsection{Cross-dataset Evaluation}
We benchmark the performance of several exiting age estimation systems under our proposed SC protocol and compare with that of our proposed one.
Rothe~\etal~\cite{Rothe2018144} proposes a classification based age estimation method wherein the labels are encoded as one-hot vectors and CE is used as the loss function for training the network.
Gao \etal models the age estimation as label distribution learning and the KL divergence is used as the loss function.
We further compare the performance with three state-of-the algorithms, \ie CE-MV~\cite{Pan20185285}, AGEn~\cite{Tan20182610} and DLDL-v2~\cite{Gao2018712}, which are the improved version of the above-mentioned methods.
We implemented all these methods.
The same model architecture (VGG), pre-processing steps and data augmentation techniques are used for a fair comparison.
In addition, the performance of the proposed scheme is compared with the available Microsoft API\footnote{\url{https://www.microsoft.com/cognitive-services/en-us/face-api/}}.

Table~\ref{tab:openset} reports the age estimation performance in terms of MAE and CS measures under the SC protocol.
As can be seen, the MAE (CS) values of the models trained using the KL loss function (\ie KL and DLDL2) are lower (higher) than those of the models trained by CE loss function (CE, Agen and CE-MV algorithms). 
This confirms the advantage of label distribution learning based methods, compared with classification based approach. 
It can be inferred from Table~\ref{tab:openset} that adopting the our proposed scheme provides a significantly higher age estimation accuracy than that achieved by the existing methods. 
The experimental results show better generalisation power of the proposed method in the cross-dataset testing, which involves unseen scenarios.
It should be emphasised that the best competing method, \ie DLDL2-v2, utilises a regularisation term which is used jointly with the KL loss function to train the model. 
Since the regularisation term affects different scales due to the existence of outliers, training the network is sensitive to the proper selection of values of regularisation parameters. 
As a positive point, our proposed loss function does not use any regularisation parameter and effectively alleviates this issue, delivering  superior performance in age prediction.

As seen in Table~\ref{tab:openset}, the performance of our system is not higher than that of DLDL-v2 in the presence of facial expressions (FACES dataset). 
On the other hand, the performance of our method is significantly better than the state-of-the-art on the other datasets, which contains of images with different ethnicity, illumination conditions, resolution and quality.
These results show our system is more robust to variations in image quality than the state-of-the-art methods. Coping with changing image quality is a challenging issue in machine learning based algorithms\footnote{In the longer version of this work~\cite{AkbariPami20}, we performed extensive experiments to show the generalisation power of our proposed loss function.}.

\subsection{Discussion}
\subsubsection{Cross-dataset vs. intra-dataset settings}
It is a well known fact that the accuracy of age estimation systems under the intra-dataset protocols is much greater than that under the cross-dataset settings.
Obviously, unseen scenarios lead to reeducation in the performance of machine learning based methods due to the uncontrolled environment being manifest in changes in lighting, variations in the subject's head pose and appearance of expression over the face, etc.
For instance, under the random-splitting protocol, the DLDL-v2~\cite{Gao2018712} algorithm and our proposed loss achieve the MAE of less than $2$ years.
However, Table~\ref{tab:openset} shows that the performance of DLDL-v2 drops to the MAE in excess of  $5$ years old under the cross-dataset setting.
This confirms the generalisation of age estimation systems cannot be truly assessed under that the intra-dataset setting.
In other words, achieving the high accuracy under the intra-dataset settings does not certainly guarantee the good performance in real-world scenarios.
This motivates us to evaluate the system accuracy under cross-dataset protocols, such as the SC protocol.
Designing a system under this challenging protocol is our ultimate goal in this paper.
As seen in Table~\ref{tab:openset}, our proposed method achieves the best MAE values and the CS scores under the SC protocol.

\subsubsection{Sensitivity to Hyper-parameters}
\begin{figure}[!t]
\centering
\includegraphics[width=0.7
\linewidth]{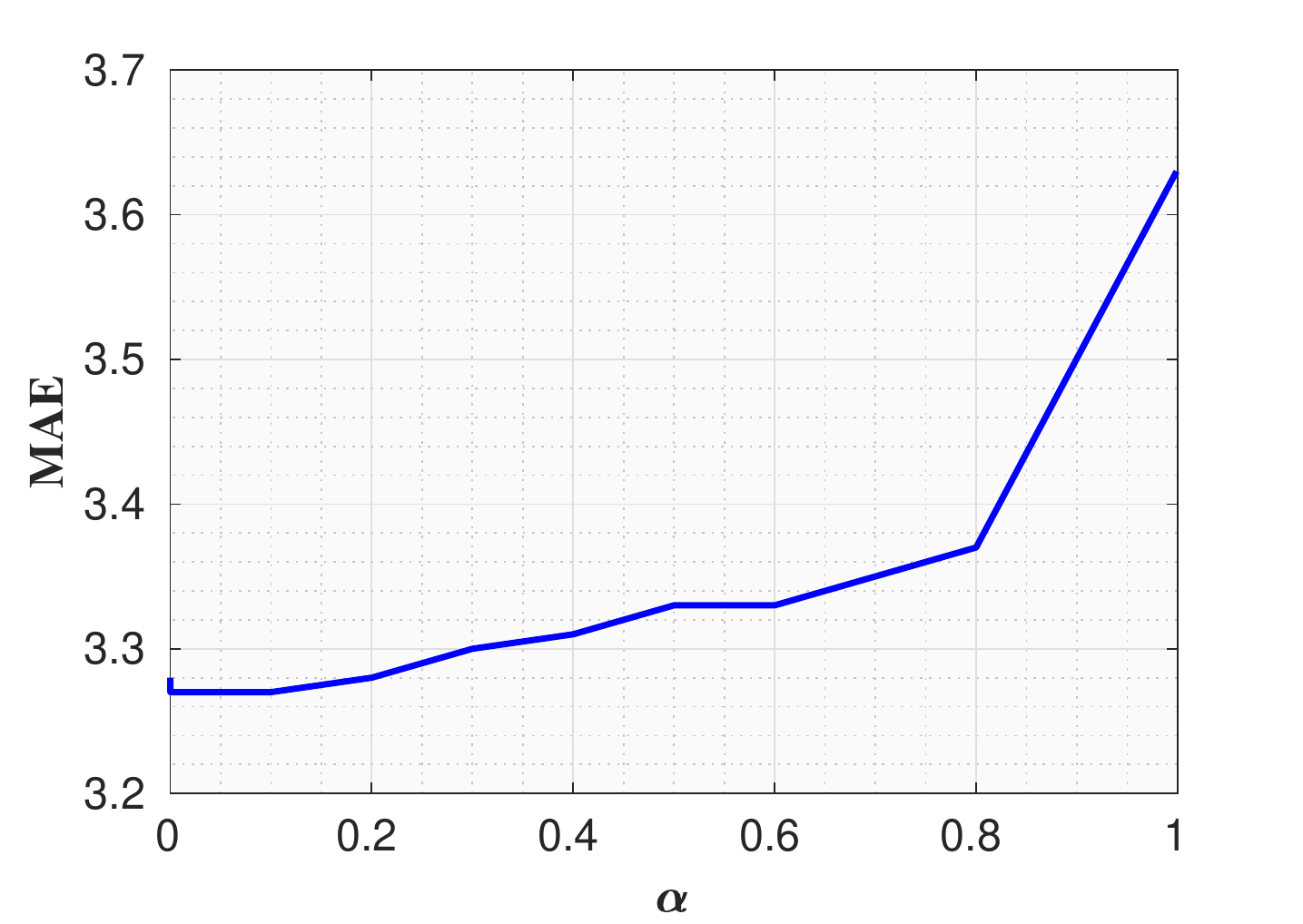}%
\caption{The effect of $\alpha$ on the age estimation accuracy.}
\label{fig:abstudy_a}
\end{figure}

We evaluate the effect of  parameter $\alpha$ on the age estimation accuracy. 
We change $\alpha$ in the range $(0,1)$ and train different models for each specific value. 
The MAE value computed  on the FG-NET dataset as a function of  $\alpha$  is shown in Fig.~\ref{fig:abstudy_a}.
Our analysis shows that the performance of age estimation is not sensitive to $\alpha$ when $\alpha$ is less than $0.2$ and the performance degrades when using large values of $\alpha$.

\section*{Acknowledgment}
This work was supported in part by the EPSRC Programme Grant (FACER2VM) EP/N007743/1 and the EPSRC/dstl/MURI project EP/R018456/1.

\bibliographystyle{ieee}
\bibliography{main}

\begin{thebibliography}{10}\itemsep=-1pt

\bibitem{AkbariPami20}
A.~{Akbari}, M.~{Awais}, Z.~{Feng}, A.~{Farooq}, and J.~{Kittler}.
\newblock Distribution cognisant loss for cross-database facial age estimation
  with sensitivity analysis.
\newblock {\em IEEE Transactions on Pattern Analysis and Machine Intelligence},
  pages 1--1, 2020.

\bibitem{Akbari20201}
A.~{Akbari}, M.~{Awais}, and J.~{Kittler}.
\newblock Sensitivity of age estimation systems to demographic factors and
  image quality: Achievements and challenges.
\newblock In {\em IEEE International Joint Conference on Biometrics (IJCB)},
  pages 1--6, 2020.

\bibitem{Akbari20161}
A.~{Akbari}, M.~{Trocan}, and B.~{Granado}.
\newblock Image error concealment using sparse representations over a trained
  dictionary.
\newblock In {\em Picture Coding Symposium (PCS)}, pages 1--5, 2016.

\bibitem{Akbari20176}
A.~{Akbari}, M.~{Trocan}, and B.~{Granado}.
\newblock Image error concealment based on joint sparse representation and
  non-local similarity.
\newblock In {\em IEEE Global Conference on Signal and Information Processing
  (GlobalSIP)}, pages 6--10, 2017.

\bibitem{Akbari20171}
A.~{Akbari}, M.~{Trocan}, and B.~{Granado}.
\newblock Joint-domain dictionary learning-based error concealment using common
  space mapping.
\newblock In {\em International Conference on Digital Signal Processing (DSP)},
  pages 1--5, 2017.

\bibitem{Akbari2559}
A.~{Akbari}, M.~{Trocan}, S.~{Sanei}, and B.~{Granado}.
\newblock Joint sparse learning with nonlocal and local image priors for image
  error concealment.
\newblock {\em IEEE Transactions on Circuits and Systems for Video Technology},
  30(8):2559--2574, 2020.

\bibitem{Angulu201842}
R.~Angulu, J.~R. Tapamo, and A.~O. Adewumi.
\newblock Age estimation via face images: a survey.
\newblock {\em EURASIP Journal on Image and Video Processing}, 2018(1):42, Jun
  2018.

\bibitem{Bashar20201}
M.~{Bashar}, A.~{Akbari}, K.~{Cumanan}, H.~Q. {Ngo}, A.~G. {Burr}, P.~{Xiao},
  and M.~{Debbah}.
\newblock Deep learning-aided finite-capacity fronthaul cell-free massive mimo
  with zero forcing.
\newblock In {\em IEEE International Conference on Communications (ICC)}, pages
  1--6, 2020.

\bibitem{Bashar20201678}
M.~{Bashar}, A.~{Akbari}, K.~{Cumanan}, H.~Q. {Ngo}, A.~G. {Burr}, P.~{Xiao},
  M.~{Debbah}, and J.~{Kittler}.
\newblock Exploiting deep learning in limited-fronthaul cell-free massive mimo
  uplink.
\newblock {\em IEEE Journal on Selected Areas in Communications (JSAC)},
  38(8):1678--1697, 2020.

\bibitem{info11020125}
A.~Buslaev, V.~I. Iglovikov, E.~Khvedchenya, A.~Parinov, M.~Druzhinin, and
  A.~A. Kalinin.
\newblock Albumentations: Fast and flexible image augmentations.
\newblock {\em Information}, 11(2), 2020.

\bibitem{Carletti2019}
V.~{Carletti}, A.~{Greco}, G.~{Percannella}, and M.~{Vento}.
\newblock Age from faces in the deep learning revolution.
\newblock {\em IEEE Transactions on Pattern Analysis and Machine Intelligence},
  pages 1--1, Apr 2019.

\bibitem{Chang2011585}
K.~{Chang}, C.~{Chen}, and Y.~{Hung}.
\newblock Ordinal hyperplanes ranker with cost sensitivities for age
  estimation.
\newblock In {\em IEEE Conference on Computer Vision and Pattern Recognition
  (CVPR)}, pages 585--592, June 2011.

\bibitem{Chen2017742}
S.~{Chen}, C.~{Zhang}, M.~{Dong}, J.~{Le}, and M.~{Rao}.
\newblock Using ranking-cnn for age estimation.
\newblock In {\em IEEE Conference on Computer Vision and Pattern Recognition
  (CVPR)}, pages 742--751, July 2017.

\bibitem{Ebner2010}
N.~C. Ebner, M.~Riediger, and U.~Lindenberger.
\newblock Faces---a database of facial expressions in young, middle-aged, and
  older women and men: Development and validation.
\newblock {\em Behavior Research Methods}, 42(1):351--362, Feb 2010.

\bibitem{Eidinger20142170}
E.~{Eidinger}, R.~{Enbar}, and T.~{Hassner}.
\newblock Age and gender estimation of unfiltered faces.
\newblock {\em IEEE Transactions on Information Forensics and Security},
  9(12):2170--2179, Dec 2014.

\bibitem{FatemifarICIP2020}
S.~{Fatemifar}, M.~{Awais}, A.~{Akbari}, and J.~{Kittler}.
\newblock A stacking ensemble for anomaly based client-specific face spoofing
  detection.
\newblock In {\em 2020 IEEE International Conference on Image Processing
  (ICIP)}, pages 1371--1375, 2020.

\bibitem{wingloss}
Z.~{Feng}, J.~{Kittler}, M.~{Awais}, P.~{Huber}, and X.~{Wu}.
\newblock Wing loss for robust facial landmark localisation with convolutional
  neural networks.
\newblock In {\em IEEE/CVF Conference on Computer Vision and Pattern
  Recognition}, pages 2235--2245, 2018.

\bibitem{Gao20172825}
B.~{Gao}, C.~{Xing}, C.~{Xie}, J.~{Wu}, and X.~{Geng}.
\newblock Deep label distribution learning with label ambiguity.
\newblock {\em IEEE Transactions on Image Processing}, 26(6):2825--2838, June
  2017.

\bibitem{Gao2018712}
B.-B. Gao, H.-Y. Zhou, J.~Wu, and X.~Geng.
\newblock Age estimation using expectation of label distribution learning.
\newblock In {\em Proceedings of International Joint Conference on Artificial
  Intelligence (IJCAI)}, pages 712--718, 7 2018.

\bibitem{Geng20161734}
X.~{Geng}.
\newblock Label distribution learning.
\newblock {\em IEEE Transactions on Knowledge and Data Engineering},
  28(7):1734--1748, July 2016.

\bibitem{Geng20132401}
X.~{Geng}, C.~{Yin}, and Z.~{Zhou}.
\newblock Facial age estimation by learning from label distributions.
\newblock {\em IEEE Transactions on Pattern Analysis and Machine Intelligence},
  35(10):2401--2412, Oct 2013.

\bibitem{Grgic2011}
M.~Grgic, K.~Delac, and S.~Grgic.
\newblock Scface -- surveillance cameras face database.
\newblock {\em Multimedia Tools and Applications}, 51(3):863--879, Feb 2011.

\bibitem{Guo2009112}
G.~{Guo}, , Y.~{Fu}, and T.~S. {Huang}.
\newblock Human age estimation using bio-inspired features.
\newblock In {\em IEEE Conference on Computer Vision and Pattern Recognition
  (CVPR)}, pages 112--119, June 2009.

\bibitem{Han20151148}
H.~{Han}, C.~{Otto}, X.~{Liu}, and A.~K. {Jain}.
\newblock Demographic estimation from face images: Human vs. machine
  performance.
\newblock {\em IEEE Transactions on Pattern Analysis and Machine Intelligence},
  37(6):1148--1161, June 2015.

\bibitem{He20173849}
Z.~{He}, X.~{Li}, Z.~{Zhang}, F.~{Wu}, X.~{Geng}, Y.~{Zhang}, M.~{Yang}, and
  Y.~{Zhuang}.
\newblock Data-dependent label distribution learning for age estimation.
\newblock {\em IEEE Transactions on Image Processing}, 26(8):3846--3858, Aug
  2017.

\bibitem{Huang20174058}
D.~{Huang}, L.~{Han}, and F.~D. l.~{Torre}.
\newblock Soft-margin mixture of regressions.
\newblock In {\em IEEE Conference on Computer Vision and Pattern Recognition
  (CVPR)}, pages 4058--4066, July 2017.

\bibitem{KhalidBHIS2020410}
S.~S. {Khalid}, M.~{Awais}, Z.~H. {Feng}, C.~H. {Chan}, A.~{Farooq},
  A.~{Akbari}, and J.~{Kittler}.
\newblock Resolution invariant face recognition using a distillation approach.
\newblock {\em IEEE Transactions on Biometrics, Behavior, and Identity
  Science}, 2(4):410--420, 2020.

\bibitem{Liu2018292}
H.~{Liu}, J.~{Lu}, J.~{Feng}, and J.~{Zhou}.
\newblock Label-sensitive deep metric learning for facial age estimation.
\newblock {\em IEEE Transactions on Information Forensics and Security},
  13(2):292--305, Feb 2018.

\bibitem{Moschoglou20171997}
S.~{Moschoglou}, A.~{Papaioannou}, C.~{Sagonas}, J.~{Deng}, I.~{Kotsia}, and
  S.~{Zafeiriou}.
\newblock Agedb: The first manually collected, in-the-wild age database.
\newblock In {\em IEEE Conference on Computer Vision and Pattern Recognition
  Workshops (CVPRW)}, pages 1997--2005, July 2017.

\bibitem{Pan20185285}
H.~{Pan}, H.~{Han}, S.~{Shan}, and X.~{Chen}.
\newblock Mean-variance loss for deep age estimation from a face.
\newblock In {\em IEEE Conference on Computer Vision and Pattern Recognition},
  pages 5285--5294, June 2018.

\bibitem{FGNET}
G.~{Panis}, A.~{Lanitis}, N.~{Tsapatsoulis}, and T.~F. {Cootes}.
\newblock Overview of research on facial ageing using the fg-net ageing
  database.
\newblock {\em IET Biometrics}, 5(2):37--46, May 2016.

\bibitem{Parkhi2015}
O.~M. Parkhi, A.~Vedaldi, and A.~Zisserman.
\newblock Deep face recognition.
\newblock In {\em British Machine Vision Conference}, 2015.

\bibitem{MORPH}
K.~{Ricanek} and T.~{Tesafaye}.
\newblock Morph: a longitudinal image database of normal adult age-progression.
\newblock In {\em International Conference on Automatic Face and Gesture
  Recognition (FGR)}, pages 341--345, Apr 2006.

\bibitem{Rothe2018144}
R.~Rothe, R.~Timofte, and L.~Van~Gool.
\newblock Deep expectation of real and apparent age from a single image without
  facial landmarks.
\newblock {\em International Journal of Computer Vision}, 126(2):144--157, Apr
  2018.

\bibitem{Seong2018TowardsFL}
S.~Seong, Y.~Lee, Y.~Kee, D.~Han, and J.~Kim.
\newblock Towards flatter loss surface via nonmonotonic learning rate
  scheduling.
\newblock In {\em The Conference on Uncertainty in Artificial Intelligence
  (UAI)}, 2018.

\bibitem{Shen20182304}
W.~{Shen}, Y.~{Guo}, Y.~{Wang}, K.~{Zhao}, B.~{Wang}, and A.~{Yuille}.
\newblock Deep regression forests for age estimation.
\newblock In {\em IEEE Conference on Computer Vision and Pattern Recognition
  (CVPR)}, pages 2304--2313, June 2018.

\bibitem{Shen2017834}
W.~Shen, K.~ZHAO, Y.~Guo, and A.~L. Yuille.
\newblock Label distribution learning forests.
\newblock In {\em Advances in Neural Information Processing Systems (NIPS)},
  pages 834--843. Curran Associates, Inc., 2017.

\bibitem{Tan20182610}
Z.~{Tan}, J.~{Wan}, Z.~{Lei}, R.~{Zhi}, G.~{Guo}, and S.~Z. {Li}.
\newblock Efficient group-n encoding and decoding for facial age estimation.
\newblock {\em IEEE Transactions on Pattern Analysis and Machine Intelligence},
  40(11):2610--2623, Nov 2018.

\bibitem{Matconvnet}
A.~Vedaldi and K.~Lenc.
\newblock Matconvnet -- convolutional neural networks for matlab.
\newblock In {\em {ACM} International Conference on Multimedia}, 2015.

\bibitem{Xu2012}
H.~Xu and S.~Mannor.
\newblock Robustness and generalization.
\newblock {\em Machine Learning}, 86(3):391--423, Mar 2012.

\bibitem{Zeng20191906}
X.~Zeng, C.~Ding, Y.~Wen, and D.~Tao.
\newblock Soft-ranking label encoding for robust facial age estimation.
\newblock {\em CoRR}, abs/1906.03625, 2019.

\bibitem{Jing2019}
J.~Zhang, W.~Li, P.~Ogunbona, and D.~Xu.
\newblock Recent advances in transfer learning for cross-dataset visual
  recognition: A problem-oriented perspective.
\newblock {\em ACM Computing Surveys}, 52(1), Feb. 2019.

\bibitem{Zhang201722492}
K.~{Zhang}, C.~{Gao}, L.~{Guo}, M.~{Sun}, X.~{Yuan}, T.~X. {Han}, Z.~{Zhao},
  and B.~{Li}.
\newblock Age group and gender estimation in the wild with deep ror
  architecture.
\newblock {\em IEEE Access}, 5:22492--22503, 2017.

\bibitem{MTCNN}
K.~{Zhang}, Z.~{Zhang}, Z.~{Li}, and Y.~{Qiao}.
\newblock Joint face detection and alignment using multitask cascaded
  convolutional networks.
\newblock {\em IEEE Signal Processing Letters}, 23(10):1499--1503, Oct 2016.

\bibitem{Zhang20174352}
Z.~{Zhang}, Y.~{Song}, and H.~{Qi}.
\newblock Age progression/regression by conditional adversarial autoencoder.
\newblock In {\em IEEE Conference on Computer Vision and Pattern Recognition
  (CVPR)}, pages 4352--4360, July 2017.

\end{thebibliography}

\end{document}